**I Code or AI Code:**
**A Comparative Evaluation of AI-rated Scores in Classroom Observations**

**Authors**

| | | |
|---|---|---|
| Yasmin Fong [1] | ykmfong@eduhk.hk | ORCID: 0000-0002-4643-1258 |
| Jane Xiang [1] | xiang@eduhk.hk | ORCID: 0009-0001-5836-7265 |
| Tak-Yue Dickson Chan [1] | ctakyue@eduhk.hk | ORCID: 0009-0003-8079-9835 |
| Kerry Lee [2*] | kerry.lee@yccece.edu.hk | ORCID: 0000-0002-8296-8028 |
| Eva Yi Hung Lau [1*] | evalau@eduhk.hk | ORCID: 0000-0002-1176-0457 |

[1] Department of Early Childhood Education, The Education University of Hong Kong, 10 Lo Ping Road, Tai Po, Hong Kong SAR, China
[2] Yew Chung College of Early Childhood Education, 2 Tin Wan Hill Road, Tin Wan, Hong Kong SAR, China

**Corresponding authors**
(*) Correspondence concerning this article should be addressed to Eva Lau (evalau@eduhk.hk), Department of Early Childhood Education, The Education University of Hong, 10 Lo Ping Road, Tai Po, Hong Kong SAR, China, and/or to Kerry Lee (kerry.lee@yccece.edu.hk), President's Office, Yew Chung College of Early Childhood Education, 2 Tin Wan Hill Road, Tin Wan, Hong Kong SAR, China.

**Acknowledgements**
This study was supported by a project grant awarded to the corresponding author from the Hong Kong Jockey Club Charities Trust, Hong Kong. We thank all the research assistants and personnel for their data collection efforts, as well as all the teachers and children who participated in this study. The views expressed in this paper are the authors' and do not necessarily represent the views of their respective institutions.

**I Code or AI Code:**
**A Comparative Evaluation of AI-rated Scores in Classroom Observations**

**Abstract**

Classroom observations are widely recognized as a key tool for establishing benchmarks of education quality and guiding pedagogical improvement, yet they remain resource-intensive and dependent on trained observers. This study evaluated the feasibility of using a LLM (GPT-5 model) to score teacher-child interactions in early childhood classrooms, benchmarked against human raters. The study analyzed 87 video-recorded observations from 38 classrooms across 30 kindergartens in Hong Kong. Using observation transcripts, the AI model was configured to apply the full Classroom Assessment Scoring System (CLASS) framework. AI-rated scores were then compared with human ratings by examining correlations and difference in mean scores of the CLASS domains and dimensions.

The results showed greater convergence between AI and raters for the Emotional Support domain and, in particular, the Quality of Feedback dimension, which captures how teachers use feedback to extend children's learning. Greater divergence emerged for interactions that were more procedural or context-dependent, particularly within the Classroom Organization and Instructional Support domains. These findings suggest that transcript-based AI scoring may capture some of the relative variation in teacher-child interactions but cannot yet reproduce calibrated human judgements consistently across the full CLASS framework. AI-assisted observation may therefore be more appropriate as a preliminary screening tool rather than as a replacement for trained observers, providing teachers with evidence for reflection rather than high-stakes evaluation. Future research should examine whether domain-specific training and incorporation of contextual and visual information can improve alignment between AI and human rated scores.

# I Code or AI Code:
# A Comparative Evaluation of AI-rated Scores in Classroom Observations

## 1. Introduction

Classroom observation is widely recognized as a critical approach for evaluating instructional quality and supporting educator professional development (e.g., Grimm et al., 2014), particularly through frameworks such as the Classroom Assessment Scoring System (CLASS; Pianta et al., 2008a), which focuses on the quality of teacher-child interactions. Despite their strong empirical grounding and widespread adoption, classroom observations are highly resource-intensive in terms of time, workload, funding, and scheduling availability (Broad, 2015). These demands are further exacerbated by substantial investments in managing rater training and calibration to ensure consistency (Matsumura et al., 2010; Cohen et al., 2003). As a result, there is a growing need for alternative or complementary approaches that can preserve the rigor of classroom observation while reducing its logistical and resource burdens.

Recent advances in AI, particularly with large language models (LLMs) models, offer promising opportunities to address these challenges. While existing research on AI in education has largely focused on efficiency for student-centered assessment tasks (e.g., Moore & Lee, 2024), comparatively little empirical work has examined the application of AI to standardized classroom observation. Addressing this gap, the present study examines the methodological applicability of AI in conducting classroom observations by comparing AI-rated and human-rated CLASS observation scores derived from videotaped classroom interactions. By analyzing both convergence and divergence in scoring patterns across domains and dimensions, this study seeks to evaluate the feasibility, reliability, and limitations of AI-assisted classroom observation, and to inform ongoing discussions about the appropriate role of AI in supporting scalable, evidence-based classroom quality assessment.

### 1.1 Value of Classroom Observations

Classroom observations are a foundational pedagogical practice for assessing and enhancing educational quality worldwide. Beyond its evaluation function, systematic observations provide educators with concrete, evidence-based data about their instructional practices, creating meaningful opportunities for reflection and professional growth (Tarusha & Bushi, 2024; Siddiqua, 2019). There is evidence that informal classroom observation approaches, such as brief, timely, and intentional visits conducted throughout the year, can be beneficial to teachers (Ing, 2010). Through these observation processes, teachers can collaborate with trained professionals and peers to reflect on authentic classroom data within their unique educational contexts (Grimm et al., 2014). Participation in such professional and peer communities enables teachers to identify patterns in their practice, pose reflective questions about areas for improvement, and implement evidence-based instructional changes (Desimone et al., 2025; Wylie et al., 2025). Over time, this cumulative reflective practice supports the development of pedagogically sound strategies that enhance student learning outcomes (Hafen et al., 2015). However, the absence of standardized procedures and observer training often introduces subjective and procedural inconsistencies between teachers and their observers (Wang & Day, 2001), underscoring the need for standardized observation protocols to improve consistency and reliability.

To address these needs, researchers have developed a range of standardized observational instruments to examine classroom quality from a more objective perspective. A prominent example is the Classroom Assessment Scoring System (CLASS; Pianta et al., 2008a), which emphasizes the quality of teacher-child interactions. Through instruments like CLASS, classroom quality can be observed through a grounded-theoretical lens, generating empirical data that can inform formative reflection and summative evaluation of educator practices, thereby supporting instructional improvement and quality assurance (Bukhalenkova et al., 2023).

In particular, the CLASS framework is grounded in developmental theories of process quality, which posit that children's learning and development are shaped most directly by the quality of their proximal interactions with adults and peers (Bronfenbrenner & Morris, 1998; Pianta et al., 2002). CLASS operationalizes this perspective by systematically assessing observable features of teacher-child interactions, including teachers' verbal instructions, emotional responsiveness, and nonverbal behaviors in everyday classroom settings. This structured approach to classroom observation has been extensively validated across diverse geographic and demographic contexts (e.g., Downer et al., 2012; Hu et al., 2016; Karuppiah, 2021; Pakarinen et al., 2010) and has been associated with teacher practices and child outcomes (Hamre et al., 2014; Hoang et al., 2019; Leyva et al., 2015). Moreover, CLASS has been widely integrated into teacher professional development and improvement initiatives, with evidence supporting its utility as a tool for reflective practice, coaching, and instructional support (e.g., Early et al., 2016; Pianta et al., 2008b; Zan & Donegan-Ritter, 2014).

**1.2 Challenges in Resource Intensity for Classroom Observations**

Despite the well-established pedagogical value of classroom observations for supporting teacher professional development and evaluating program quality, their implementation poses substantial resource challenges for many education organizations. These challenges situate classroom observations within a broader set of constraints affecting teacher professional development, including time pressures, increased teacher workloads, limited funding, and restricted scheduling availability (e.g., Broad, 2015; Fairman et al., 2022), all of which may limit their prioritization in practice. As a result, although classroom observation is widely endorsed as evidence-based practice, its routine and sustained implementation are often difficult to prioritize in school settings.

These logistical constraints may be resolved through video-based observation as an alternative to live observation (e.g., Borg, 2021). Compared with in-person observations, video recordings allow for classroom practices and dynamics to be captured remotely and reviewed multiple times, providing flexibility and analytic depth for observers (Ault et al., 2019). Video-based approaches may also reduce teachers' anxiety associated with the physical presence of an observer, thereby supporting more naturalistic instructional behaviors and partially mitigating the Hawthorne effect, whereby teachers alter their practice when being observed (Mackey, 2017; Page, 2016). As such, video observations offer a less intrusive means of examining classroom processes and may yield more authentic representations of everyday instructional practices (Curby et al., 2016).

However, video-based observation does not eliminate challenges associated with human raters. Observer fatigue and bias remain difficult to avoid, particularly in large-scale studies requiring prolonged coding sessions and repeated reliability checks. Prior research has demonstrated that fatigue-related drift can introduce increased variability and reduce measurement precision over time (Matsumura et al., 2010; Raudenbush et al., 2007). These cognitive and analytic demands continue to contribute to the overall resource burden of classroom observation when data collection is conducted remotely. Beyond fatigue, extensive rater calibration and training are required to ensure reliable and valid observation coding. Validated instruments, such as the CLASS measure, necessitate intensive initial training and ongoing reliability monitoring to maintain measurement validity (Pianta et al., 2008a; Hamre et al., 2013). This requirement reflects the inherently interpretative nature of classroom observation, which demands fine-grained judgments about complex and dynamic instructional interactions, often under time constraints (Bell et al., 2012). Sustaining reliability over time presents additional challenges, particularly in large-scale projects involving multiple raters, where continuous calibration and recalibration are necessary to maintain scoring consistency (Allen et al., 2013). These processes substantially increase the human, temporal, and financial resources required for classroom observations.

Collectively, the demands associated with observation logistics, observer fatigue, intensive training, and calibration intensify the cumulative resource burden of classroom observation. These persistent challenges raise concerns about the long-term feasibility and scalability of traditional observational approaches, particularly in resource-constrained educational contexts, and underscore the need for complementary solutions that support reliable, efficient, and sustainable classroom-quality assessment.

### 1.3 Growing prominence of AI as a resource

In recent years, AI has advanced rapidly, driven in part by the emergence of LLMs that enable increasingly sophisticated computational learning and human-like interaction (Kasneci et al., 2023). These developments have expanded AI's capacity to simulate aspects of human learning, comprehension, problem-solving, decision-making, and creativity (e.g., Duan et al., 2019; Topol, 2019), positioning AI as a transformative technology with growing potential to alleviate resource constraints across sectors. In education, AI-powered systems, particularly conversational and generative tools, have gained prominence for their ability to support real-time problem-solving, generate high-quality instructional content, personalize learning processes, and provide rapid feedback to students and educators (e.g., Bettayeb et al., 2024). Consequently, much of the existing research has focused on technology-assisted teaching and learning, and on educational assessment practices, where AI has demonstrated considerable benefits such as increased efficiency, automated grading, feedback generation, and adaptive instructional support (Chen et al., 2022; Chu et al., 2025; Huang et al., 2023; Owan et al., 2023). As a result, AI and their LLM have been widely used to support student-centered evaluation activities, including assignment analysis, formative feedback, and reporting, thereby reducing educators' workload (e.g., Liao et al., 2024; Liebenow et al., 2025).

Despite their growing prominence, the application of AI in education has remained largely confined to task-based evaluations, with comparatively limited attention given to how AI-powered tools may be used in classroom evaluations to inform educators' instructional practices or professional development. One early attempt to address this gap was conducted by Shapsough and Zualkernan (2018), who examined the automation of classroom observation by testing machine-learning algorithms within a mobile application. Using the Stallings Classroom Snapshot observation system (Stallings, 1977), their approach achieved 68.9% accuracy rate in labeling classroom activities from audio recordings. While these findings demonstrated the technical feasibility of automating certain aspects of classroom observation, the reported accuracy was relatively modest and achieved by only one of the 10 machine algorithms tested, highlighting both the promise and the limitations of early technology-based observational systems. Other work has approached AI-assisted observation from a different angle. For example, Singh et al. (2025) examined verbal exchanges between teachers and students in lecture settings, using AI to detect nuanced patterns in speech and verbal dynamics. Although this innovative work offered valuable insights into factors associated with lecture quality, it remained primarily focused on audio-based interactions and did not fully capture the broader dynamics of classroom settings, such as social relationships, curriculum structure, and instructional elements.

Addressing the need for broader applicability, more recent work has moved beyond single-exchange analysis toward scaling AI-assisted observation across entire classroom sessions. A recent pilot study by Li et al. (2026) explored the use of AI as a 'teammate' for scaling up classroom observations. Drawing on multiple LLM agents, the study developed a pipeline that transcribed classroom audio, contextualized the language, and evaluated it using an observational rubric. This approach achieved an 18-fold efficiency gain compared with traditional classroom observation methods involving human raters. While this was a notable advancement, the rubric was limited to binary coding (present/not present) across two subscales from the ECQRS-EC and SSTEW frameworks, which may constrain its broader applicability in capturing the complexity of classroom interactions. Collectively, these emerging studies demonstrate the growing potential of AI to support classroom evaluation. However, these studies remain constrained by limited scope, fragmented capture of interactions, and simplified rubric systems that warrant further empirical investigation.

### 1.4 The Current Study

The present study contributes to the growing discourse on AI in education by examining the feasibility and reliability of LLM for scoring classroom observations. Classroom observation is a data-intensive process that depends heavily on trained human observers, which can limit the scalability of evidence-based classroom quality assessment in educational settings. In response to these challenges, this study investigates whether AI can perform the complex interpretive work traditionally undertaken by trained human observers when evaluating instructional interactions during classroom observation. Guided by this aim, the study addressed the following research question: To what extent do AI-rated scores align with human raters' scores in classroom observations?

To answer this question, the CLASS measure was selected as the observational framework, given its status as a well-validated and standardized instrument for assessing classroom quality through measuring teacher-child interactions in early childhood settings (Hamre et al., 2014). Drawing on prior exploratory studies (Shapsough & Zualkernan, 2018; Singh et al., 2025), we hypothesized that AI could rate observation scores closely aligned with those of human raters in instructional language exchanges (i.e., convergence in Instructional Support and its dimensions). Given that the interactions within the Emotional Support and Classroom Organization domains, and their corresponding dimensions (Pianta et al., 2008a), may only be partially represented in transcripts, we were unsure of the alignment between AI and raters.

## 2. Method

### 2.1 Procedures

The ethics review committee from the corresponding authors' university approved the current study before any data was obtained. Teachers, corresponding children, and parents of the respective kindergartens provided written consent for videotaping classroom activities throughout the children's attendance on a typical kindergarten day. Participants who were filmed incidentally without consent were removed in post-production, and activities that would violate participants' privacy and protection (e.g., washroom breaks) were not filmed. Classroom observations were filmed by trained research assistants who adhered to a strict and standardized videotaping protocol.

This study is part of a large-scale longitudinal project that focuses on community-based interventions to bridge developmental gaps among young children aged 3 to 6. The present study obtained video data from classroom observations conducted during the baseline year in 62 Hong Kong kindergartens, from August 2023 through the first year of kindergarten (K1), for children aged 3-4. Filming began at the start of each class session, when children entered the classroom, and continued through the end of the day's curriculum, depending on the preschool program and the participating classrooms' schedules. The recordings captured a range of typical curriculum activities, including activity corners, gross motor play, themed learning tasks, structured instructional lessons (e.g., language learning), and routine activities (e.g., snacks/mealtimes or transition to washroom breaks).

**2.2 Sample**

This study drew subsample video recordings from 30 kindergartens, representing a range of annual tuition fees: low (government-subsidized or below HKD9,999, $n = 16$); mid-range (HKD10,000-19,999, $n = 7$); and high (HKD20,000 and above, $n = 7$). The sample was selected through a screening process based on predetermined criteria for audio-visual quality and coding feasibility. Videos were purposefully selected based on the criteria that classroom exchanges were clearly captured as videos were to be audio transcribed, classrooms with a single educator were prioritized over those with multiple educators to minimize overlapping speech and enhance the clarity of observational targets (excluding 83.24% of videos).

In alignment with the CLASS framework (Pianta et al., 2008a), footage depicting children's recess or unstructured free-play periods was excluded to maintain consistency with the measure's coding requirements. Based on this screening process, only videos that met all the quality and relevance criteria were selected for subsequent coding and analysis. The finalized sample included video recordings from 38 K1 classroom teachers (47.4% female, 52.6% missing; average teaching experience of 10.89 years, $SD = 7.42$), totaling approximately 26.5 hours for analysis. The high percentage of missing demographic data was attributable to teacher non-reporting (see Table 1 for more details), though these variables are not considered central to the study. In total, there were 87 classroom observations, each lasting 15-20 minutes, with an average of two per classroom to capture variability across classroom activities.

**Table 1**

*Teacher demographics*

| | *n* | *Mean* | *SD* |
|---|---|---|---|
| Gender | | | |
| Female | 18 | | |
| Missing | 20 | | |
| Age (years) | 18 | 32.17 | 7.42 |
| Missing | 20 | | |
| Teaching experience (years) | | 10.89 | 7.99 |
| Missing | 20 | | |
| Education attainment | 18 | | |
| Up to Diploma Certificate | 2 | | |
| Bachelor's Degree | 12 | | |
| Postgraduate or Master's Degree | 3 | | |
| Other Qualifications | 1 | | |
| Missing | 20 | | |

**2.3 Measures**

***2.3.1 The Classroom Assessment Scoring System Pre-K***

The Classroom Assessment Scoring System Pre-K (CLASS Pre-K; Pianta et al., 2008a) was used as the classroom observation measure to evaluate classroom quality. The CLASS Pre-K framework comprises three overarching domains derived from observed dimensions, as shown in Figure 1. The Emotional Climate domain captured the warmth, responsiveness, and emotional tone in teacher-child relationships that help children feel safe to participate in the classroom through its four corresponding dimensions (Positive Climate, Negative Climate, Teacher Sensitivity, and Regard for Student Perspectives). The Classroom Organization domain focuses on how well the teacher structures behavior, time, and engagement to ensure learning runs smoothly through three corresponding dimensions (Behavior Management, Productivity, and Instructional Learning Formats). The Instructional Support domain examines how teachers extend thinking, provide feedback, and build on children's vocabulary through meaningful interactions through three corresponding dimensions (Concept Development, Quality of Feedback, and Language Modeling). The domains provided a broad overview of teacher-child interaction quality, while a focus on the constituent dimensions allows for a more detailed examination of specific interactional practices. Domain scores were aggregated from individual dimension scores, which were rated from a 7-point Likert scale from (1) low to (7) high quality.

**Figure 1**

*CLASS Scale Overview*

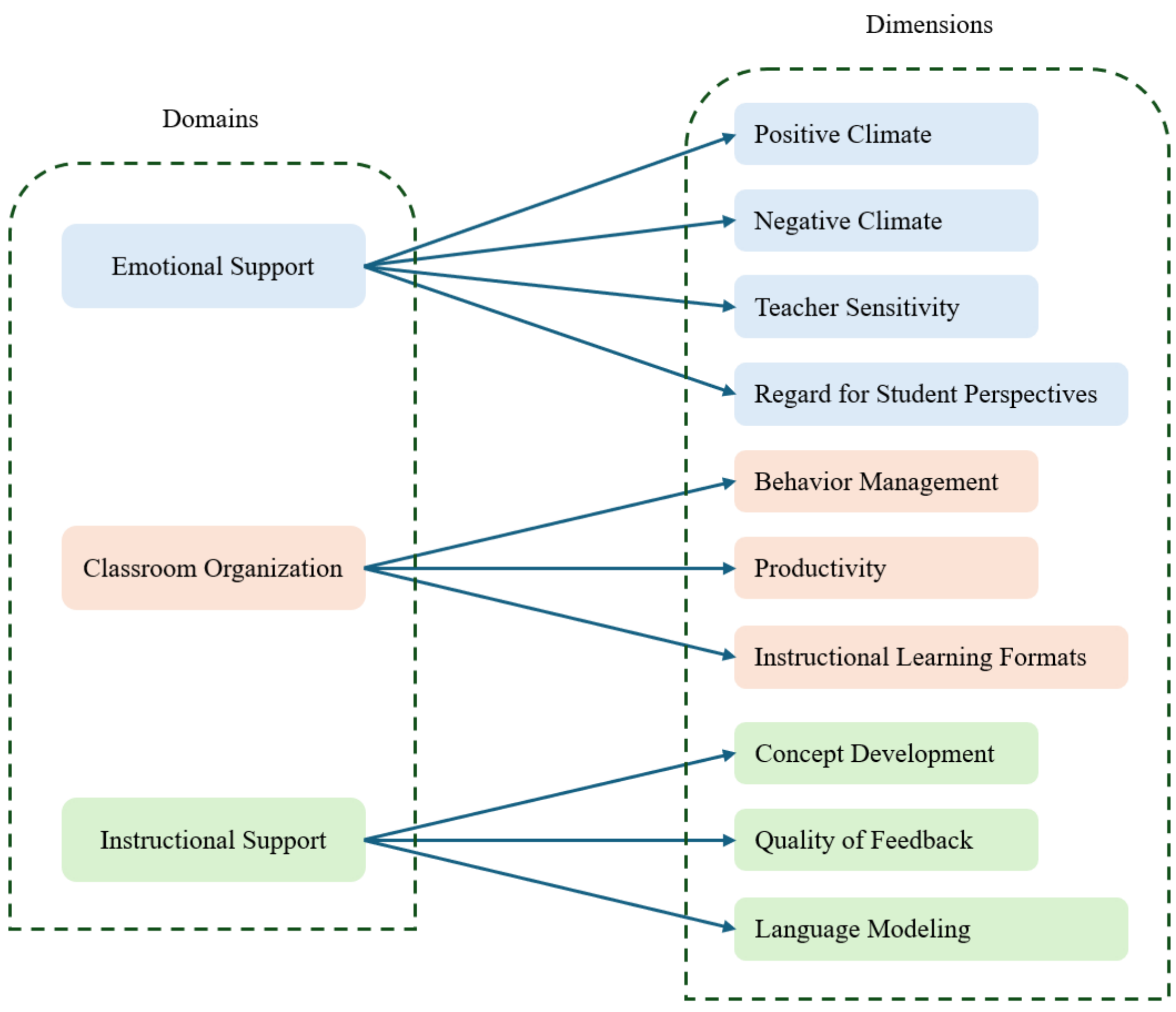

Typically, the CLASS methodology requires between four and seven observations per classroom, which are averaged to provide a more representative estimate of overall educator-child interaction quality. This approach is primarily intended to evaluate classroom-level quality. However, as the primary aim of the present study was to examine the applicability and consistency of raters and AI during the coding process, ratings from individual observation cycles were compared to more precisely evaluate similarities and differences between raters and AI-rated coding.

### *2.3.2 Human raters*

In the current study, the term 'raters' refers specifically to the performance and scoring of a group of eight human raters. All raters held at least a Bachelor's Degree or were qualified early childhood education practitioners. These human raters completed a comprehensive training session on the CLASS framework led by a licensed and experienced CLASS Pre-K trainer. This comprehensive training covered the CLASS framework and observation logistics and included exemplar practice videos scored against the framework manual. Following the training, the trainer led a series of calibration rounds to ensure alignment with Hong Kong's local context and to establish calibration and inter-rater reliability standards.

The calibration process began with a first round in which all raters independently coded five identical classroom observation videos using the CLASS manual, followed by individualized feedback and a trainer-led group discussion to promote alignment in scoring. Raters then completed a second round of independent coding of five identical video observations and again received individualized feedback based on their scores. After achieving satisfactory alignment, the remaining 77 classroom observation videos were coded independently, with each observation randomly assigned to two different raters to ensure ongoing cross-checking of scores. Using dimension scores, the interrater reliability among the human raters was examined with weighted Cohen's kappa ($\kappa w$), yielding values ranging from .672 to .861, indicating moderate to substantial agreement with a certified trainer. While these values may be lower than typical thresholds, earlier studies of CLASS have noted minimum criteria of .60 (La Paro et al., 2004; Sandilos & DiPerna, 2011). Subsequently, a consolidated score, averaged across overlapping raters, was calculated for each observation and used for analysis.

### *2.3.3 AI Rater*

Within the scope of this study, 'AI' or 'AI model' specifically refers to the performance and scoring rated by the Perplexity GPT-5.0 model (Perplexity, 2025), which served as the LLM for this comparative analysis. All AI training and coding occurred on August 7, 2025, 8:00 PM to August 8, 5:30 AM Pacific Time (PT), using the most up-to-date artificial intelligence model at the time of analysis. Given the difficulty of AI models in accurately interpreting video content at the time of this study, all classroom observation videos were transcribed for AI training and coding. Classroom observation transcriptions were completed by fluent or native bilingual (Chinese and English) research personnel who were not part of the CLASS coding process. Transcriptions were completed based on the classroom medium of instruction and the interaction between teachers and children, which was predominantly in Chinese.

The AI model was trained using a prompt-engineering approach, a critical component for optimizing LLM learning and performance. Initial training prompts were developed by identifying essential instructions and requirements that guided how the model should learn and utilize the CLASS framework for scoring. These prompts were subsequently refined through interactive interactions with the AI model to improve clarity and effectiveness (see the detailed prompts in Appendix A). To parallel the training provided to human raters, the CLASS framework manual was provided to the AI model to establish an understanding of the observational framework and scoring procedures.

Following this, a series of structured training and calibration cycles was conducted using classroom observation notes and consolidated scores rated by the human raters. Each training cycle consisted of three steps (see Appendix B for more details): (i) a case study phase, in which the AI model was presented with two classroom observation transcripts, along with the human raters' consolidated scores and observation notes for alignment; (ii) a practice phase, during with the AI independently coded two new observation transcripts; and (iii) a feedback phase, in which consolidated scores and observations notes were shared with the AI to support reflective learning. This three-step cycle was repeated four times to strengthen the AI's interpretability of the CLASS measure and alignment to the local context. Figure 2 summarizes this training procedure.

Following completion of training and calibration, the AI model independently scored the remaining 71 classroom observation transcripts and rated scores for the respective CLASS dimensions for subsequent analysis.

**Figure 2**

*Training and calibration procedure for Humans and AI*

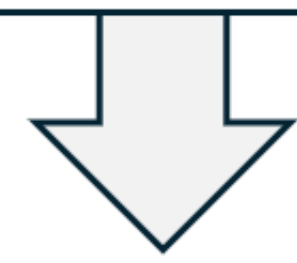

### 2.4 Data Analysis

Before conducting analysis, all Negative Climate dimension scores were reversed (i.e., higher scores indicated a lower level of negativity in the observed classroom). Outliers in the CLASS dimension scores accounted for 0.423% of the data points for raters and 0.704% for AI, which were subsequently winsorized by replacing the corresponding data points with ± 3 *SD* values for each dimension. Domain scores were then aggregated from their corresponding dimensions.

Descriptive statistics were first computed to examine the scoring of both raters and AI at the CLASS domain and dimension levels. To examine alignment between human-rated scores and AI-rated scores, correlation analyses were conducted to assess convergence or divergence in scoring patterns across the domains and dimensions. Inter-rater reliability analysis was then used to evaluate overall agreement, followed by independent *t*-tests to examine mean differences between AI and raters' scores at both the domain and dimension levels. All quantitative analyses were conducted using IBM SPSS Statistics (Version 30). Our study examined how AI-rated scores compared with those of consolidated human raters in assessing classroom observations.

## 3. Results

### 3.1 Preliminary Results

Table 2 presents the descriptive statistics for raters' and AI-rated scores across all domains and dimensions. For the Emotional Support domain, raters tended to score higher at both the domain and dimension levels than AI ratings of the corresponding domain scores and respective dimensions, with larger differences observed for the Positive Climate and Negative Climate dimensions. In the Classroom Organization domain, raters scored slightly lower than AI ratings. In general, their dimension scores were similar with minor differences, while the Behavior Management dimension showed the largest difference. Within the Instructional Support domain, raters consistently scored lower than AI ratings at both the domain and dimension level, with the Concept Development dimension showing the largest difference. Furthermore, the distributions of all measured variables were evaluated for normality by inspecting the unstandardized skewness (absolute value < 2) and kurtosis (absolute values < 3). Only the raters' Negative Climate dimension score showed a non-normal distribution, with extreme kurtosis.

**Table 2**

*Descriptive statistics of raters (n = 71) and AI (n = 71)*

| | Raters | | | | AI | | | |
|---|---|---|---|---|---|---|---|---|
| | *Mean* | *SD* | Skewness | Kurtosis | *Mean* | *SD* | Skewness | Kurtosis |
| CLASS Domains | | | | | | | | |
| Emotional Support | 5.490 | .562 | .124 | -.234 | 4.685 | .632 | -.626 | .215 |
| Classroom Organization | 5.121 | .462 | -.412 | .264 | 5.134 | .567 | -.478 | .240 |
| Instructional Support | 2.937 | .878 | .681 | -.123 | 3.221 | .799 | .093 | -.670 |
| CLASS Dimensions | | | | | | | | |
| Positive Climate | 5.435 | .793 | -.736 | 1.461 | 4.280 | .721 | -.955 | 1.229 |
| Negative Climate[1] | 6.956 | .131 | -3.034 | 8.293 | 5.870 | .559 | -.551 | 1.585 |
| Teacher Sensitivity | 5.178 | .753 | -.124 | -.353 | 4.890 | .838 | -.533 | -.075 |
| Regard for Student Perspectives | 4.385 | 1.039 | -.047 | -.579 | 3.690 | .821 | -.160 | -.437 |
| Behavior Management | 5.308 | .639 | -.441 | -.036 | 5.450 | .672 | -1.125 | 1.334 |
| Productivity | 5.338 | .764 | -.861 | 1.413 | 5.300 | .705 | -.743 | .342 |
| Instructional Learning Formats | 4.713 | .987 | -.445 | .393 | 4.650 | .927 | -.559 | .496 |
| Concept Development | 2.425 | 1.072 | .686 | -.493 | 2.830 | 1.00 | .262 | -.703 |
| Quality of Feedback | 3.316 | .957 | .632 | .383 | 3.510 | .892 | .164 | -.110 |
| Language Modelling | 3.069 | .859 | .105 | -.375 | 3.320 | .770 | -.250 | .408 |

*Note*. [1]Reversed Score.

Figure 3 illustrates the score distribution of Negative Climate ratings for both AI and raters. Given that Negative Climate was a reverse-scored dimension in which higher scores indicated lower levels of negativity, raters' observation scores were heavily concentrated at the scale ceiling, with 7 ($n$ = 63) as the most common score, followed by scores between 6 and 7 ($n$ = 8). This indicated raters generally perceived minimal negativity across observed classrooms. In contrast, AI-rated scores showed a slightly larger distributional variance, with scores predominantly at 6 ($n$ = 51), followed by 5 ($n$ = 13), 7 ($n$ = 6), and 4 ($n$ = 1).

**Figure 3**

*Distribution of Negative Climate in Raters' and AI-rated scores*

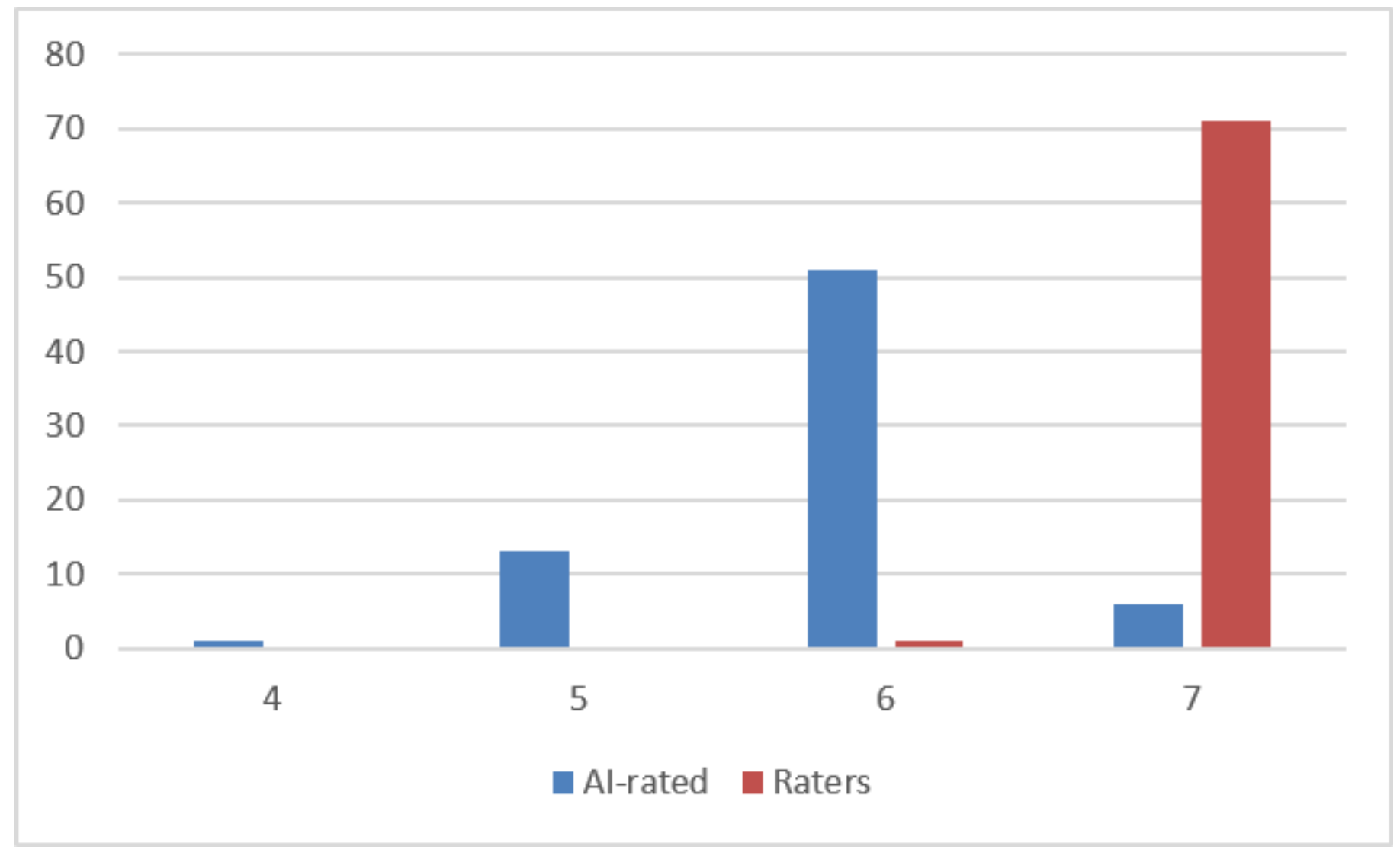

### 3.2 Scoring alignment between raters and AI

Correlation analysis was conducted to assess the associations in scores between raters and AI across the three domains and ten dimensions, as shown in Table 3. Pearson correlations were conducted on the domains, and to address the nonparametric assumptions, Spearman correlations were conducted for the dimensions.

**Table 3**

*Correlations between raters' and AI-rated scores*

| CLASS Framework | Correlations[1] ($r$) |
|---|---|
| Domains | |
| Emotional Support | .406*** |
| Classroom Organization | .123 |
| Instructional Support | .195 |
| | |
| Dimensions | |
| Positive Climate | .438*** |
| Negative Climate[2] | .085 |
| Teacher Sensitivity | .313** |
| Regard for Student Perspectives | .296* |
| Behavior Management | .161 |
| Productivity | .029 |
| Instructional Learning Formats | .136 |
| Concept Development | .132 |
| Quality of Feedback | .269* |
| Language Modelling | .144 |

*Note*. [1]Pearson correlations for domains and Spearman correlations for the dimensions. [2]Reversed Score. Significant values: *$p < .05$, **$p < .01$, ***$p < .001$.

At the domain level, a statistically significant positive correlation emerged between raters and AI on Emotional Support, indicating a modest degree of convergence. The correlations for Classroom Organization and Instructional Support were not statistically significant. At the dimension level, three of the four Emotional Support dimensions yielded significant associations. Positive Climate showed the strongest alignment, followed by Teacher Sensitivity and Regard for Student Perspectives, while the correlation for Negative Climate was not significant. None of the three dimensions within the Classroom Organization reached significance. The Quality of Feedback dimension was the only one to show a positive, significant correlation, while the remaining Instructional Support dimensions were non-significant.

Results from the inter-rater reliability between the consolidated raters' and AI-rated scores for the CLASS dimensions indicated moderate agreement, with a weighted Cohen's kappa ($\kappa w$) of 0.681 ([0.646, 0.716], $p < .001$). Based on the CLASS measurement protocol, this reported statistic would be considered a reliable standard, with dimensions agreeing to within 1 point (corresponding to $\kappa w \approx 0.60–0.75$). Furthermore, examination of individual raters' scores and AI-rated scores continued to indicate somewhat moderate agreement, with weighted Cohen's kappa (κw) values ranging from .515 to .712 (ps < .001). Subsequent analyses examined further comparisons of the domain and dimension scores in detail.

Table 4 presents the results of the independent *t*-tests comparing raters' and AI-rated scores at the domain- and dimension-levels of the CLASS framework. At the domain level, significant differences were observed: raters tended to score higher in Emotional Support and lower in Instructional Support than AI. No significant difference was found for the Classroom Organization domain.

**Table 4**

*Results from the independent t-tests*

| CLASS framework | Mean Difference | *t*-value | Degrees of freedom | *p* | Cohen's *d* |
|---|---|---|---|---|---|
| Domain Level | | | | | |
| Emotional Support | .806 | 8.002 | 140 | < .001 | 1.343 |
| Classroom Organization | -.013 | -.111 | 140 | .912 | .019 |
| Instructional Support | -.284 | -2.041 | 140 | .043 | .343 |
| | | | | | |
| Dimension Level | | | | | |
| Positive Climate | 1.152 | 9.9075 | 140 | < .001 | 1.523 |
| Negative Climate[1] | 1.084 | 16.279 | 78.204[#] | < .001 | 2.732 |
| Teacher Sensitivity | .291 | 2.178 | 140 | .031 | .365 |
| Regard for Student Perspectives | .694 | 4.419 | 140 | < .001 | .742 |
| Behavior Management | -.149 | -1.372 | 140 | .172 | .230 |
| Productivity | .041 | .360 | 140 | .720 | .060 |
| Instructional Learning Formats | .069 | .439 | 140 | .662 | .074 |
| Concept Development | -.406 | -2.336 | 140 | .021 | .392 |
| Quality of Feedback | -.191 | -1.229 | 140 | .221 | .206 |
| Language Modelling | -.255 | -1.928 | 140 | .056 | .324 |

*Note.* [1]Reversed Score. Data are presented as mean (*SD*). Mean differences refer to the raters' scores minus the AI's scores. [#]Degrees of freedom adjusted due to violation of unequal variances.

At the dimension level, significant differences between raters and AI were observed across multiple dimensions. Across the dimensions of Positive Climate, Negative Climate, Teacher Sensitivity, and Regard for Student Perspectives, raters scored significantly higher than AI. This would suggest, at the mean level, a divergence of the Emotional Support dimensions. No significant differences were observed for Behavior Management, Productivity, and Instructional Learning Formats, suggesting no observed mean difference between raters and AI in the Classroom Organization dimensions. On the other hand, significant differences were observed in the Concept Development, but not in the Quality of Feedback and Language Modeling dimensions. This suggests that a divergence within specific Instructional Support dimensions, in which raters scored significantly lower than AI. Overall, these findings suggest that there may be a domain- and dimension-specific pattern of divergence between raters and that AI-rated scoring.

## 4. Discussion

The results presented above suggest that AI-rated scoring partially aligned with human raters on the CLASS framework, but was domain-dependent and falls short of what independent classroom observation would require. Overall, the agreement was moderate, consistent with early reports on weighted kappa agreement (La Paro et al., 2004). AI-rated scores were typically within 1-scale point of raters' scores, a level that the CLASS protocol treats as an acceptable standard of reliability for observational assessment (Pianta et al., 2008a; Sandilos & DiPerna, 2011). Moderate agreement of this kind is sufficient for observational applications when the goal is to support system-level monitoring or formative feedback rather than individualized evaluation of classroom quality (Landis & Koch, 1977; Cicchetti, 1994). Our findings partially confirm and partially refute our hypothesis, and this pattern of convergence and divergence defines the current ceiling of AI-assisted observation.

### 4.1 Where AI aligns with raters: verbal and structured

Consistent with our hypothesis, convergence was strongest where classroom quality was expressed through explicit and structured language, aligning with prior studies (Shapsough & Zualkernan, 2018; Singh et al., 2025). The strength of the reported correlations was comparable to earlier findings in which multimodal AI models approximated human inter-rater reliability (Hou et al., 2024) suggesting promising consistency between human and AI evaluations. In the Quality of Feedback dimension, correlations were significant and mean differences were not, indicating minimal systematic bias. This is consistent with Whitehill and LoCasale-Crouch (2024), who reported that LLM-based predictions approximated human ratings in instructions-based support in the classroom when the construct operationalized through explicitly verbal cues and with emerging evidence of AI-human convergence in feedback evaluations (e.g., Barri & Hikmawan, 2026). The structured, exchange-based indicators of this dimension (e.g., scaffolding, contingent responses) appeared to match the pattern-recognition strengths of LLMs trained on large-scale text data (Mirchandani et al., 2023; Speer et al., 2026). This finding suggests that AI shows promising convergence in evaluating teacher-child interactions related to contingent responding and may serve as an informative indicator for teachers to reflect on their instructional practice.

Convergence was also evident in the Emotional Support domain and its dimensions with significant correlations, a finding that exceeded our initial uncertainty about this domain. The convergence suggests that AI captured the warm, positive verbal teacher-child interactions that underpin a safe and warm classroom environment (Hamre & Pianta, 2005). Yet, raters consistently assigned higher average scores. This difference likely reflected raters' sensitivity to nuanced nonverbal and relational cues, such as facial expressions and tone (Calvo & D'Mello, 2010; Pianta et al., 2012) that were not accessible through transcripts. However, it does not preclude the possibility of rater bias (Park et al., 2015). Such bias may arise from prior expectations, subjective interpretations of emotional displays, or a tendency to overvalue relational warmth in teaching contexts, thereby inflating observed differences. Despite relying primarily on transcripts, AI still achieved consistency with raters, but the persistent gap marks the limit of transcript-based scoring for this domain.

### 4.2 Where AI diverged: nonverbal, procedural, and contextual constructs

Divergence was most pronounced in the Classroom Organization and its dimensions, indicating minimal evidence of alignment between raters and AI. This aligned with our expectation of limited evidence for this domain and suggests that classroom management poses challenges for automated observation. This difficulty likely stems from inherently rapid, context-sensitive, and nonverbal aspects of classroom management practices (Li, 2006) that would challenge transcript-based LLMs (Pang et al., 2023), particularly without representation of spatial dynamics and movement of teachers and children within the classroom environment (Elbaum et al., 2024). Accounting for these nuanced cues in future LLM development may enable AI systems to better capture the complexity of classroom management, thereby enhancing the validity of automated observations and aligning them more closely with the multifaceted changes that occur in classroom environments.

The findings for Instructional Support were more inconclusive than our hypothesis anticipated. The Concept Development and Language Modeling dimensions, which rely heavily on explicit instructional language to capture pedagogical interactions (Hu et al., 2016), revealed low correlations between raters and AI. This misalignment with AI consistently having scored higher than raters on both dimensions suggests that AI may have identified additional, or incorrect, discourse features of instructional exchanges that raters may have overlooked, especially when simultaneously evaluating multiple dimensions of teaching quality (Cohen & Goldhaber, 2016). While this interpretation aligns with prior studies demonstrating of the early capabilities of LLMs in detecting teacher discourse (Kelly et al., 2018) and supports the view that classroom transcripts can provide sufficient evidence to estimate global scores for instruction-focused exchanges (Whitehill & LoCasale-Crouch, 2024), potential of incorrect lens when observing discourse evidence. This is interesting, given Quality of Feedback showed stronger convergence than the other two Instructional Support dimensions could be due to these exchanges, in which the latter dimensions requiring pedagogical focused strategies. This is consistent with evidence that well-trained AI models falter when evaluating classroom dialogue beyond their training scope (Wang et al, 2024), and points to the needed for targeted pedagogical training before AI can parallel human observers in recognizing nuanced variations in teaching practices (e.g., Bell et al., 2019). The partial support for our hypothesis is therefore instructive, indicating AI convergence where constructs are explicit and structured, but not across the Instructional Support as a whole.

Against this backdrop, these findings reveal a specific structural pattern, that partially supported the hypothesis of convergence between raters and AI in instructional language exchanges (Instructional Support). On the other hand, the unspecified hypothesis on the convergence of Emotional Support and Classroom Organization, yielded surprising results. As agreement with human raters remains partial and conditional on the construct being assessed, the present findings do not support using AI as an independent classroom observation system but support the AI-assisted observation as complementary.

However, effective implementation requires further preparation. Teachers and observers need further AI literacy training as part of professional development for classroom observations (Riggs, 2025), while LLMs require targeted pedagogical training and calibration approximate human observation (Wang & Chen, 2025). AI-assistance retains a clear value in relieving the workload of classroom observation, which may also help mitigate occurrences of rater fatigue when prolonged observation periods, repeated exposure to similar classroom interactions, and sustained cognitive load diminish observers' attentional accuracy and consistency over time (Matsumura et al., 2010; Raudenbush et al., 2007). This positions AI to serve as a first pass scoring and evidence-surfacing tool that directs human attention where it is most needed. More importantly, our findings do not imply that AI should replace authentic judgment but rather support the direction of a hybrid observation model in which AI could enhance the efficacy, consistency, and scalability, while raters retain a central role in interpreting emotional nuance and supporting professional learning. A hybrid model could offer direct implications for how schools deploy AI-assisted observation, how teachers are prepared to use it, and how future research should be prioritized.

## 5. Limitations and Future Directions

While our study provided innovative contributions on AI for classroom observations, several limitations should be acknowledged, along with directions for future research. First, due to the geographic and access constraints associated with AI tools at the time of data collection, the study examined only one LLM (i.e., Perplexity GPT-5.0) using transcript-based classroom observations. While this approach allowed for an initial examination of AI feasibility in classroom observation, reliance on a single model may limit the generalizability of the findings. Future studies should incorporate comparisons across multiple LLMs to enable more nuanced assessments of model performance and to determine whether observed patterns are model-specific or generalizable across AI platforms. Moreover, recent advances in multimodal LLM, such as Google Gemini (Gemini Team, 2023), demonstrate increasing capacity to process video and audiovisual data. This suggests promising directions for AI-assisted classroom observation for continuous investigations.

Secondly, the study's contextual scope was limited to early childhood classrooms in Hong Kong serving children aged 3-4 years. This developmental and cultural context may place greater emphasis on certain aspects of teacher-child interactions, which could have influenced the scoring patterns of both artificial and authentic raters. As a result, the findings may not generalize to classroom observations serving older students, where instructional interactions may be more cognitively and academically oriented (e.g., Tengberg et al., 2022). Future research should extend this line of inquiry to primary and secondary education settings to test the adaptability and effectiveness of AI systems across a broader range of instructional contexts, age groups, and pedagogical demands. Moreover, our analysis sample was limited to a single teacher per video due to transcription constraints, thereby excluding over 80% of the full video sample. Although this reduced the representativeness of the sample and may have limited the capture of classroom interaction variability, we do not expect it to substantially affect our findings, as the study focused on the accuracy of AI-rated coding, rather than the assessment of teachers' classroom quality. Nevertheless, these factors may limit the generalizability of our findings. Future research could address this by developing scalable pipelines, similar to those described by Li et al. (2026), to support large-scale LLM-based observations.

Third, the present study relied exclusively on the CLASS measure as the observational framework for comparison and used single observations. While CLASS is a well-validated and widely used instrument, using a single scale and limited observation rounds may limit the interpretability and stability of the findings. We recommend that future studies increase the number of observational cycles to expand sample size and power for multilevel analyses, enabling more robust examinations of reliability patterns and consistency in classroom observations. Additionally, future research should also consider alternative observational frameworks, such as the Early Childhood Environment Rating Scale (ECERS; Harms et al., 1998) for assessing structural enablers in the classroom, or the International Comparative Analysis of Learning and Teaching (ICALT; van der Lans et al., 2018) for assessing teachers' learning instruction in older grade classrooms. Doing so would examine whether the performance is consistent across different conceptualizations of classroom quality and observational protocols.

## 6. Conclusions and Implications

This study evaluated the methodological applicability of AI to support classroom observation by comparing AI-rated scores with those of trained human raters using the full CLASS observational framework. The findings showed that AI can approximate human judgment of an acceptable reliability for formative and system-level purposes, but only in a specific CLASS dimension (Quality of Feedback) and domain (Emotional Support). Divergence was evident when observations depended on nonverbal, procedural, and contextual cues (Classroom Organization and Instructional Support domain). Our findings conclude that AI-assisted observation can serve as a limited complementary to human judgement.

From a practical perspective, the findings suggest that schools with limited human resources may leverage AI as a reference or screening tool for classroom evaluation, particularly when capacity for intensive human observation is constrained. AI-assisted approaches can help alleviate resource demands in domains such as Emotional Support while still informing instructional improvement, and they may be particularly helpful for teachers who want to reflect on their feedback and prompting practices, indicators central to the Quality of Feedback. For these benefits to be realized, AI-assisted observation should be positioned as a complement to, not a replacement for, human raters, and teacher-facing tools should be introduced with explicit training on the model's capabilities and limits. AI outputs should be framed as evidence for reflection rather than as scores for evaluation. More broadly, the integration of AI into classroom observation systems holds promise for providing scalable, cost-effective monitoring of classroom quality, offering policymakers and educational leaders new opportunities to support professional development and system-level improvement, particularly where transcript-based approaches can substitute for audiovisual data raised from data-protection and ethical concerns.

From a research perspective, the findings highlight the need for further studies to compare multiple AI models, incorporate multimodal data sources (e.g., video), and examine diverse educational contexts to strengthen the validity and generalizability of AI-assisted observation. Research should also evaluate whether AI-assisted workflows genuinely reduce rater workload without compromising validity, and whether targeted training of LLMs on early childhood classroom discourse closes the alignment gap in instruction-focused dimensions. Continued collaboration between educational researchers and AI developers will be essential to optimize AI systems for classroom observation and to ensure their responsible, ethical, and effective use in supporting teaching and learning. Moving forward, research should not be framing whether AI can replace classroom observation, but how AI-assisted observations can be designed so that teachers can retain judgement where it matters and freed from workload where AI can demonstrate reliability.